\documentclass[sigconf]{acmart}

\usepackage{algorithm}
\usepackage{xspace}
\usepackage{algorithmic}
\usepackage[inline]{enumitem}
\usepackage{multirow}
\usepackage{subcaption}
\usepackage{cleveref}

\setcopyright{none}
\renewcommand\footnotetextcopyrightpermission[1]{}

\begin{document}
\title{Mimicking Human Visual Development for Learning Robust Image Representations}

\author{Ankita Raj}
\email{ankita.raj@cse.iitd.ac.in}
\orcid{0000-0003-1068-9406}
\affiliation{
	\institution{Indian Institute of Technology Delhi}
	\city{New Delhi}
	\country{India}}

\author{Kaashika Prajaapat}
\email{cs1170337@gmail.com}
\orcid{0009-0007-7228-7206}
\affiliation{
	\institution{Clinikally}
	\city{Gurugram}
	\country{India}}

\author{Tapan Gandhi}
\email{Tapan.Kumar.Gandhi@ee.iitd.ac.in}
\orcid{0000-0002-3532-9389}
\affiliation{
	\institution{Indian Institute of Technology Delhi}
	\city{New Delhi}
	\country{India}}

\author{Chetan Arora}
\email{chetan@cse.iitd.ac.in}
\orcid{0000-0003-0155-0250}
\affiliation{
	\institution{Indian Institute of Technology Delhi}
	\city{New Delhi}
	\country{India}}

%\renewcommand{\shortauthors}{Raj et al.}

\vspace{2em}	
\begin{abstract}
The human visual system is remarkably adept at adapting to changes in the input distribution—a capability modern convolutional neural networks (CNNs) still struggle to match. Drawing inspiration from the developmental trajectory of human vision, we propose a progressive blurring curriculum to improve the generalization and robustness of CNNs. Human infants are born with poor visual acuity, gradually refining their ability to perceive fine details. Mimicking this process, we begin training CNNs on highly blurred images during the initial epochs and progressively reduce the blur as training advances. This approach encourages the network to prioritize global structures over high-frequency artifacts, improving robustness against distribution shifts and noisy inputs.
Challenging prior claims that blurring in the initial training epochs imposes a stimulus deficit and irreversibly harms model performance, we reveal that early-stage blurring enhances generalization with minimal impact on in-domain accuracy. Our experiments demonstrate that the proposed curriculum reduces mean corruption error (mCE) by up to 8.30\% on CIFAR-10-C and 4.43\% on ImageNet-100-C datasets, compared to standard training without blurring. Unlike static blur-based augmentation, which applies blurred images randomly throughout training, our method follows a structured progression, yielding consistent gains across various datasets. Furthermore, our approach complements other augmentation techniques, such as CutMix and MixUp, and enhances both natural and adversarial robustness against common attack methods. Code is available at \url{https://github.com/rajankita/Visual_Acuity_Curriculum}.
\end{abstract}

%\begin{CCSXML}
%	<ccs2012>
%	<concept>
%	<concept_id>10010147.10010257</concept_id>
%	<concept_desc>Computing methodologies~Machine learning</concept_desc>
%	<concept_significance>300</concept_significance>
%	</concept>
%	<concept>
%	<concept_id>10010147.10010257.10010293.10010294</concept_id>
%	<concept_desc>Computing methodologies~Neural networks</concept_desc>
%	<concept_significance>300</concept_significance>
%	</concept>
%	<concept>
%	<concept_id>10010147.10010178.10010224</concept_id>
%	<concept_desc>Computing methodologies~Computer vision</concept_desc>
%	<concept_significance>500</concept_significance>
%	</concept>
%	</ccs2012>
%\end{CCSXML}
%
%\ccsdesc[300]{Computing methodologies~Machine learning}
%\ccsdesc[300]{Computing methodologies~Neural networks}
%\ccsdesc[500]{Computing methodologies~Computer vision}

\keywords{Generalization, Robustness, Deep neural networks, Representation learning.}

\maketitle
\pagestyle{plain}

\section{Introduction}

\begin{figure}[t]
	\centering
	\vspace{2em}	
	\includegraphics[width=\linewidth]{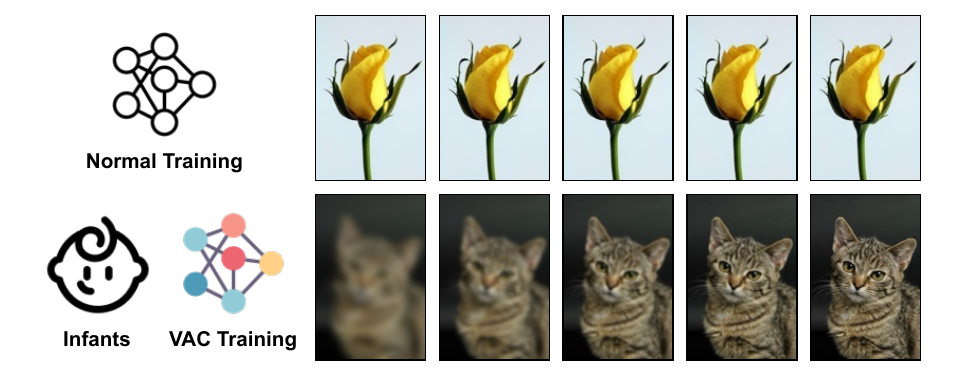}
	\caption{CNNs are trained using high resolution images from the first epoch (top row). Newborn babies, on the other hand, start with poor visual acuity and gradually develop sharper vision as their visual system matures. We propose a training curriculum, VAC, that starts with highly-blurred training images and gradually improves resolution (bottom row). }
	\vspace{1em}	
\end{figure}

Convolutional Neural Networks (CNNs) have achieved remarkable performance in computer vision tasks in recent years, even surpassing human performance on certain tasks. However, despite their remarkable success \cite{krizhevsky2012imagenet,ren2015faster}, CNNs suffer from a large drop in performance when the test distribution differs from the training distribution \cite{li2017deeper}. CNNs are highly sensitive to natural image corruptions such as noise and blur \cite{geirhos2018generalisation}, as well as to carefully crafted adversarial perturbations \cite{goodfellow2014explaining}, limiting their reliability in real-world applications. 
In contrast, human vision is not easily disrupted by such shifts in the input distribution, suggesting fundamental differences in how humans and machines learn visual representations. This paper takes a step towards bridging this gap by identifying one such difference rooted in the developmental trajectory of human visual perception. 

A growing body of work addresses CNN robustness to distribution shifts through data augmentation \cite{cubuk2019autoaugment,cubuk2020randaugment}, where augmented images generated by applying low-level distortions to the original data are included during training.
Techniques such as AugMix \cite{hendrycks2019augmix}, CutMix \cite{yun2019cutmix}, and DeepAugment \cite{hendrycks2021many} employ diverse and aggressive train-time augmentations to improve robustness to natural image corruptions.
While these methods have proven effective, they typically apply augmentations uniformly throughout training, without accounting for the evolving learning capacity of the model. This static treatment of input complexity may limit the model’s ability to learn robust representations. Furthermore, like most work in representation learning, these techniques primarily focus on asymptotic performance in the later stages of training, overlooking the importance of early learning dynamics in shaping model robustness. 

In this work, we take cues from the developmental trajectory of the human visual system to design a learning strategy for CNNs aimed at fostering robust representations. In humans and other animals, visual acuity is initially poor at birth and gradually improves over the first few months of life \cite{burkhalter1993development,siu2018development}. Neuroscientific studies suggest that initial poor retinal acuity promotes extended spatial processing in the visual cortex, which is critical for tasks such as configural face analysis \cite{vogelsang2018potential}. Interestingly, Vogelsang \emph{et~al.} \cite{vogelsang2018potential} observed similar effects in CNNs: training face recognition models with blurred images in the early epochs increased the effective receptive field and improved recognition performance. We extend this idea beyond face recognition and propose a generalized curriculum—Visual Acuity Learning (\texttt{VAC})—that progressively transitions CNNs from low to high frequency visual input during training.

In \texttt{VAC}, training begins with highly blurred inputs, and the blur level is progressively reduced as training advances, ultimately reaching unaltered high-resolution images. This structured curriculum encourages the model to first attend to low-frequency, global structures before being exposed to high-frequency details. As a result, the network is less prone to overfitting on texture-level noise and spurious high-frequency artifacts. To mitigate the risk of catastrophic forgetting of the representations learned from earlier blur levels, we incorporate a replay mechanism that periodically reintroduces previous blur stages during later training phases. This mechanism not only preserves robustness acquired in earlier phases but also supports smoother adaptation to increasing image detail.

It is important to note that the proposed Visual Acuity Curriculum (\texttt{VAC}) differs critically from Gaussian blur augmentation. In blur augmentation, clean and blurred images are typically intermixed throughout training, exposing the network to a full spectrum of frequency information at all times. This approach lacks temporal structure and does not account for the evolving learning capacity of the model. In contrast, \texttt{VAC} introduces a deliberate information deficit during the early stages of training by presenting only heavily blurred inputs. This controlled progression gradually increases the level of visual detail over time. The initial period of information deficit is a key factor in our approach, as it compels the network to focus on global, low-frequency features rather than memorizing fine-grained noise. This structured exposure leads to better generalization across distribution shifts compared to static data augmentation strategies.

The proposed method offers a somewhat contrasting view to the work of Achille \emph{et~al.} \cite{achille2018critical}, who argue that introducing blur in the early stages of training of a neural network induces low-level deficits, resulting in an irreversible loss in accuracy. Their work draws a parallel to the \textit{critical learning period} observed in animals, where sensory deprivation during early development can cause permanent impairments in specific visual functions \cite{hubel1970period,konishi1985birdsong}. We contend that this critical learning period hypothesis is incomplete. In contrast to their findings, we show that blurring in the early training stages encourages the model to learn robust, low-frequency representations, thereby improving generalization with only a marginal reduction in in-domain accuracy. For many real-world applications, this trade-off is not only acceptable, but desirable, as it yields models that are significantly more reliable in the face of real-world noise and corruption.

\vspace{1em}
\noindent\textbf{Contributions:} We make the following contributions:
\begin{enumerate}
	\item We propose \texttt{VAC}, a training curriculum inspired by human visual development where images are blurred in the initial epochs and progressively restored to their original form.
	\item Through comprehensive experiments across multiple datasets and network architectures, we demonstrate that \texttt{VAC} improves generalization under real-world dataset shifts — evaluated using standard benchmarks for natural robustness — outperforming vanilla training and other curriculum learning methods like CBS \cite{sinha2020curriculum} and FixRes \cite{touvron2019fixing}.
	\item We show that \texttt{VAC} is compatible with popular data augmentation techniques such as MixUp \cite{zhang2018mixup} and $\ell_2$-adversarial training \cite{madry2018towards}, providing further improvements in generalization.
	\item Overall, we provide a holistic perspective on the impact of blurring-based deficits in early training epochs. While the critical learning period hypothesis \cite{achille2018critical} holds so far as the in-domain accuracy is concerned, we argue that such deficits are not inherently detrimental. In fact, the marginal drop in in-domain accuracy is outweighed by a significant boost in robustness to distribution shifts.
\end{enumerate}
\section{Related Work}

\subsection{Human Visual Development and Critical Learning Period} 
Humans and several other animals are born with immature visual functions due to structural immaturities in the visual pathway \cite{burkhalter1993development,siu2018development}. 
Contrast sensitivity \cite{atkinson1977contrast, pirchio1978infant},  visual acuity \cite{hamer1989development}, orientation selectivity \cite{braddick1986orientation, morrone1986evidence}, binocular vision \cite{birch1996fpl, birch1982stereoacuity} and several other visual capabilities progressively improve at different rates following birth. Among these, visual acuity, which is the ability to resolve fine spatial detail, is particularly underdeveloped at birth. Newborn humans typically exhibit acuity below 20/600 \cite{vogelsang2018potential, sokol1978measurement}, which gradually improves to the standard adult acuity of 20/20 over the first few years.

Vogelsang \emph{et~al.} \cite{vogelsang2018potential} hypothesize that this early period of poor acuity may not simply be a developmental limitation, but rather a functional feature that promotes the emergence of spatially extended receptive fields in the visual cortex. The reduced visual detail in early sensory input encourages spatial integration over larger regions of the visual field, helping the brain form robust, global representations of objects. 

On the other hand, a few studies have shown that several visual capabilities have associated periods of enhanced plasticity, termed as `critical' periods, sensory deprivation during which negatively impacts normal visual development. Following the famous experiment by Weisel and Hubel \cite{wiesel1963effects, hubel1970period} where the eyes of kittens were sutured to study critical periods for ocular dominance, critical periods have been observed for song learning in birds \cite{konishi1985birdsong}, and amblyopia\cite{taylor1979critical} and stereopsis \cite{fawcett2005critical} in humans among other tasks. Analogous studies have been done for artificial neural networks as well, where researchers have shown that initial training phase plays a significant role in the performance of CNNs \cite{frankle2020early}. In their critical learning period theory for CNNs, Achille \emph{et~al.} \cite{achille2018critical} argue that stimulus deficit during the initial epochs, such as that induced by blurring of training data, results in an irrevocable loss in a network's performance. Other works show how the optimization trajectory of a CNN is affected by the hyperparameters of Stochastic Gradient Descent \cite{jastrzebski2019break}, learning rate \cite{jastrzebski2021catastrophic} or regularization \cite{golatkartime}. 

We believe that the findings of \cite{achille2018critical} and \cite{vogelsang2018potential}, as described above, may not be contradictory to each other. However, each may be incomplete in the sense that, whereas continuous sensory deprivation may affect accuracy, initial deprivation followed by improvement in visual acuity may be necessary to learn robust representations. 

\subsection{Robustness in Convolutional Neural Networks}
CNNs, while excelling at large-scale vision tasks, are sensitive to input distortions. The study of robustness in CNNs covers both common image corruptions and worst-case adversarial perturbations. 

Common image corruptions include diverse image degradations like noise, rain, fog, snow, and blur, etc. \cite{recht2019imagenet,geirhos2018generalisation,hendrycks2018benchmarking,ovadia2019can}, and is typically evaluated using the ImageNet-C \cite{hendrycks2018benchmarking} dataset. Geirhos \emph{et~al.} \cite{geirhos2018generalisation} pointed out that robustness to image distortions can be achieved by training on distorted images, yet it fails to generalize to previously unseen distortion types. Other methods for improving robustness to out-of-distribution data include data augmentation \cite{yin2019fourier,hendrycks2021many,hendrycks2019augmix,rusak2020simple}, increasing model size, reducing texture-bias \cite{geirhos2018imagenet}, pre-training on huge datasets \cite{hendrycks2019using}, as well as combinations of these \cite{hendrycks2021many}.  We show that \texttt{VAC} improves robustness to image distortions without being trained on distorted images.  \texttt{VAC} can also be used in conjunction with these methods to further improve robustness. 

Adversarial noise refers to small, imperceptible perturbations crafted using methods like FGSM \cite{goodfellow2014explaining} and PGD \cite{madry2018towards} that can drastically degrade CNN performance. Adversarial robustness techniques often rely on Adversarial Training \cite{madry2018towards} or advanced versions like TRADES \cite{zhang2019theoretically}, FSR \cite{kim2023feature}, etc. 

Most existing robustness techniques apply transformations statically throughout training. In contrast, our method progressively adjusts the input resolution to guide the model toward robust feature extraction. 

\subsection{Curriculum Learning and Training Schedules}

Curriculum learning \cite{bengio2009curriculum} suggests that models benefit from training on easier examples first, gradually increasing task complexity. Curriculum learning by both gradually increasing the difficulty of data \cite{pmlr-v97-hacohen19a,pmlr-v70-graves17a}, or the modeling capacity of the network \cite{croitoru2022lerac}, has been shown to improve convergence as well as performance. 

Our proposed training technique can be seen as a specific instance of curriculum learning. Different from the existing approaches, in this work, we define a curriculum in terms of a blur schedule. Sinha \emph{et~al.} \cite{sinha2020curriculum} proposed a curriculum by blurring the activation maps of convolutional layers. Unlike \cite{sinha2020curriculum}, we perform blurring of input images rather than feature maps, and are the first ones to study the impact of blurring-based deficit on generalization. Burduja \emph{et~al.} \cite{burduja2021unsupervised} proposed gradually deblurring input images to improve 3D medical image registration. Blurred images are easier to align compared to sharper images, thus their curriculum proceeds from easier to gradually more difficult concepts. On the other hand, classification using blurred images is more difficult. Thus gradually decreasing input blur has significantly different connotations for image classification, which we address in this work.
Touvron \emph{et~al.} \cite{touvron2019fixing} discussed the impact of the resolution of training images on test accuracy. They suggested that training with lower resolution images, and further finetuning with higher resolution images improves accuracy. Unlike \cite{touvron2019fixing}, the proposed curriculum trains on gradually increasing image resolutions, and does not require additional fine-tuning on higher resolution images.

\section{Proposed Method}
\begin{figure*}[t]
	\centering
	\includegraphics*[width=\linewidth]{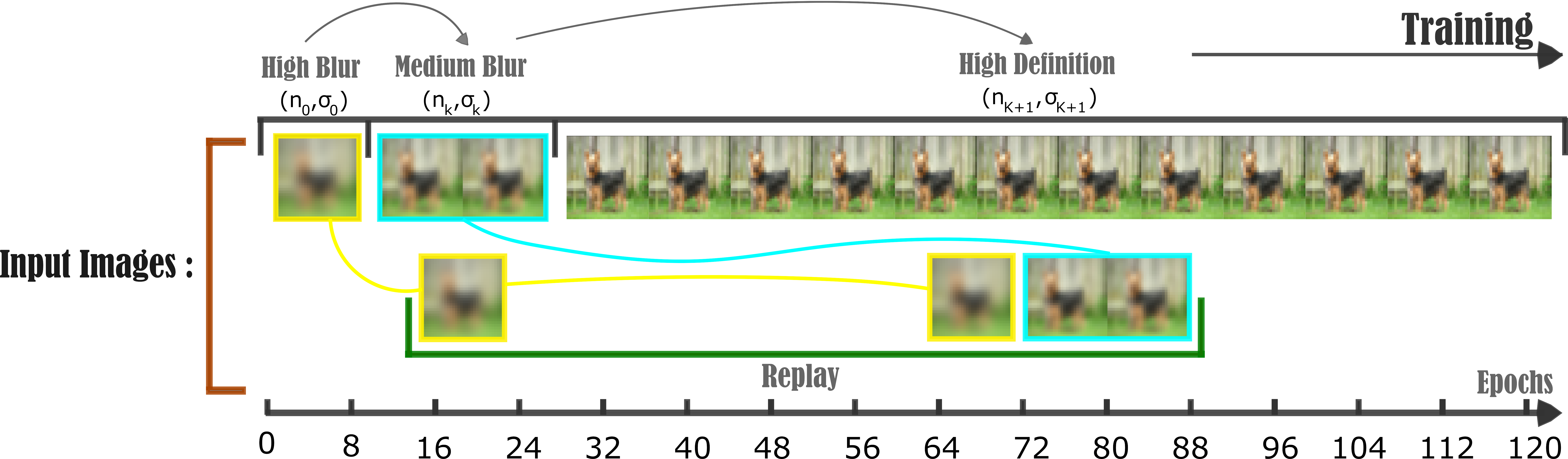}
	\caption{Visual representation of the proposed training curriculum. The entire training routine is divided into segments, each segment characterized by number of epochs $n_k$, and standard deviation $\sigma_k$. A segment $(n_k, \sigma_k)$ includes images blurred with a Gaussian blur kernel with standard deviation $\sigma_k$ as well as a replay of previously seen blur levels $\sigma_{<k}$. Training starts with highly blurred images and progresses towards sharper images. }
	\label{fig:method}
	\Description{Visual representation of the varying levels of blur in training images across epochs.}
\end{figure*}

Visual acuity in newborn humans and other animals develops gradually after birth, whereas CNNs are exposed to high resolution images from the very first epoch. Inspired by the findings of \cite{vogelsang2018potential}, we propose Visual Acuity Curriculum--a training routine that mimics the development of visual acuity in newborns over the first few months. 

\subsection{Progressive Blurring Curriculum}
To simulate varying levels of visual acuity, we apply Gaussian Blur to the training images, following \cite{vogelsang2018potential}. Prior works suggest that Gaussian blur suppresses high-frequency, texture-related components, encouraging models to focus on more global, shape-driven cues \cite{geirhos2018imagenet,yin2019fourier}. 

Let $x \in \mathbb{R}^{H \times W \times 3}$ be a training image. The blurred version of the image is defined as:
\begin{equation}
	\tilde{x} = \mathcal{G}_\sigma * x,
\end{equation}
where $\mathcal{G}_\sigma$ denotes a Gaussian kernel with standard deviation $\sigma$, and $*$ denotes convolution. A larger $\sigma$ results in stronger blur, effectively removing high-frequency information and simulating lower visual acuity.

The training process is divided into a sequence of $K$ curriculum segments. Each segment $k \in \{1, \dots, K\}$ is parameterized by a tuple $(n_k, \sigma_k)$, where $n_k$ is the number of epochs and $\sigma_k$ is the blur level used in that segment. To simulate the gradual improvement in acuity, we start with a high value of $\sigma_{max}$ and halve $\sigma$ every few epochs (refer \Cref{alg1}). We choose $\sigma_\textrm{max}$ based on the image resolution of the training dataset, with higher $\sigma$ for larger images. For convenience, $\sigma_\textrm{max}$ is taken to be a power of 2. The number of epochs per segment is designed to grow exponentially, loosely mirroring the early rapid development and later gradual refinement in acuity observed in human vision \cite{chandna1991natural}. {We use a deficit only for the initial $20\%$ epochs}. 

As an example, for the CIFAR-10 dataset with image resolution $32 \times 32$, we train for 200 epochs starting with an initial $\sigma_\textrm{max} = 2$, with the schedule:
\begin{equation}
	\{ (13, 2),\ (27, 1),\ (160, 0) \},
\end{equation}
In this curriculum, the first segment of $13$ epochs uses $\sigma=2$, second segment of $27$ epochs corresponds to $\sigma=1$, and finally for the remaining $160$ epochs $\sigma=0$ is used which corresponds to using the original images (without blur) from the dataset. 

\begin{algorithm}[t]
	\caption{Define curriculum}\label{alg1}
	\begin{algorithmic}[1]
		\REQUIRE { Total epochs $N$, highest blur $\sigma_\textrm{max}$}
		\ENSURE{ \{($n_k$, $\sigma_k$ )\}, epochs and blur for segments of curriculum}
		\STATE Deficit epochs, $N_\textrm{def} = \lfloor N/5 \rfloor$\ ($\lfloor \cdot \rfloor$ denotes the floor operation)
		\STATE $K = \log_2 (\sigma_\textrm{max})$\;
		\STATE $\sigma_0 = \sigma_\textrm{max}$
		\STATE $n_0 = {N_\textrm{def}}/{\left(\sum_{i=0}^{K}2^i\right)}$
		\FOR{$k = \{1, \dots K\}$}
		\STATE $\sigma_k = \sigma_{k-1}/2$\;
		\STATE $n_k = n_{k-1}\times2$\;
		\ENDFOR
		\STATE $\sigma_{K+1} = 0$\;
		\STATE $n_{K+1} = N -  N_\textrm{def}$\;
	\end{algorithmic}
\end{algorithm}

\begin{algorithm}[t]
	\caption{Train using \texttt{VAC}}\label{alg2}
	\begin{algorithmic}[1]
		\REQUIRE{Training data $D^\textrm{train}$, Curriculum $\{(n_k, \sigma_k )\}_{k=0}^{K+1}$}
		\ENSURE{Trained model with parameters $W^{*}$}
		\STATE Initialize model parameters to $W$\;
		\FOR{$\textrm{epoch} = \{1, \sum_{k=0}^{K+1}{n_k} \}$}
		\STATE $i \gets$  index of current curriculum segment \quad \COMMENT{\textbf{Line 1}}
		\FOR{$(x,y) \in D^\textrm{train}$}
		\STATE $j \gets$ sample from $\{0, \dots i\}$  \quad \COMMENT{\textbf{Line 2}}
		\STATE $\tilde{x} = \mathcal{G}_{\sigma_j} * x$  \quad \COMMENT{\textbf{Line 3}} 
		\STATE $\mathcal{L} \gets \textrm{loss}(\tilde{x}, y) $\;
		\STATE $W \gets \textrm{update}(W, \mathcal{L})$\;
		\ENDFOR
		\ENDFOR
		\STATE \textbf{Return} {$W$}   
	\end{algorithmic}
\end{algorithm}

\subsection{Blur Replay}
While the progressive blurring schedule allows the model to transition from low to high-frequency information, we observed a sharp degradation in performance on blurred inputs after transitioning to clean images. This is a classic symptom of catastrophic forgetting \cite{kirkpatrick2017overcoming}, wherein representations acquired in earlier training phases are overwritten by new updates.

To address this, we introduce a replay mechanism inspired by the continual learning concepts of rehearsal and memory replay that mitigate catastrophic forgetting \cite{kirkpatrick2017overcoming}. Specifically, we inject a fraction of training examples processed using earlier blur levels into later segments. During the initial training segment, all images are blurred using a Gaussian kernel with the maximum blur level, $\sigma_\textrm{max}$. In subsequent segments, to prevent catastrophic forgetting of early-stage representations, we introduce a data replay mechanism. Specifically, each training image is blurred using a $\sigma$ value sampled from all previously encountered blur levels. Let $n_j$ denote the number of epochs for blur level $\sigma_j$. The sampling probability for blur level $\sigma_j$ in segment $k$ is given by:
\begin{equation}
	p_j = \frac{n_j}{\sum_{i=1}^k n_i}, \quad \text{for } j \leq k.
\end{equation}

This ensures that blur levels encountered earlier in the curriculum continue to be revisited proportionally in later stages, thereby reinforcing previously learned representations. As a result, the network learns to maintain robustness across a range of visual conditions rather than specializing only for high-resolution inputs. The training process for the proposed Visual Acuity Curriculum is outlined in \Cref{alg2}. 

\subsection{Implementation Simplicity}
One of the practical strengths of VAC is its ease of implementation. The entire training routine can be realized with minimal changes to a standard training loop, specifically, by dynamically adjusting the blur level for each training image using a schedule and introducing a weighted sampler during the later stages. In our implementation, this requires only three additional lines of code compared to standard network training (these modifications are highlighted in \Cref{alg2}), making VAC compatible with most existing training pipelines.

\section{Experiments and Results} 

\subsection{Experiment Setup}

\begin{figure}[t]
	\centering
	\includegraphics*[width=0.95\linewidth]{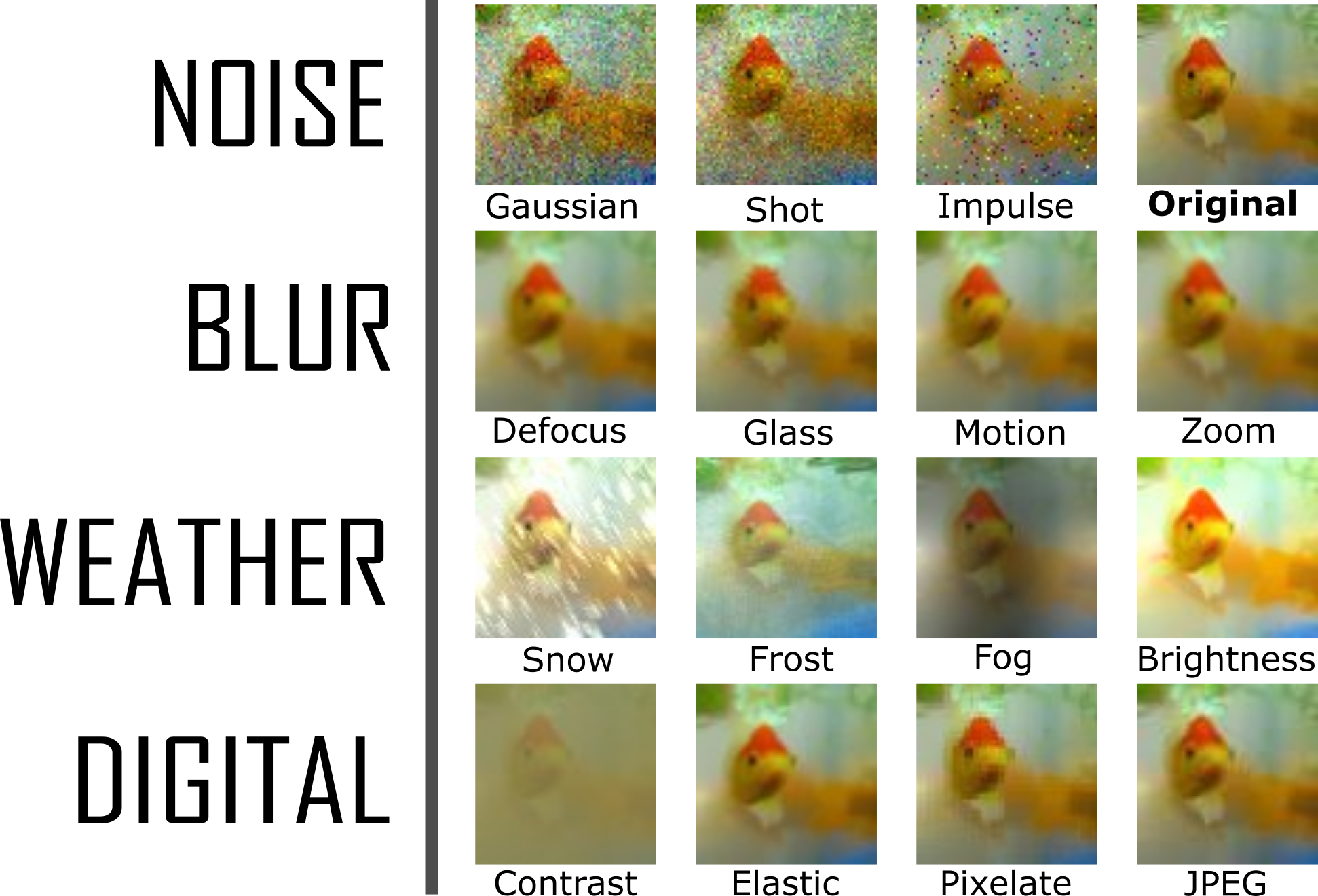}
	\caption{Examples of different image corruptions from the ImageNet-100-C dataset.}
	\Description{Different types of image corruptions visualized on an image of a goldfish.}
	\label{fig:corruption_images}
\end{figure}

\subsubsection{Datasets} We use two RGB image datasets for our experiments. {CIFAR-10} dataset \cite{krizhevsky2009learning} comprises of $32\times32$ low-resolution images belonging to 10 object classes, with $50,000$ training and $10,1000$ testing images. {ImageNet-100} \cite{laidlaw2020perceptual} is a larger dataset  with $224 \times 224$ images, which is a subset of the original ImageNet dataset \cite{ILSVRC15}, obtained by sampling every tenth class from the original dataset.  

\subsubsection{Corruption Datasets} To evaluate performance under distribution shift, we use common corruption benchmarks \cite{hendrycks2018benchmarking}: CIFAR-10-C, and ImageNet-100-C. These are corrupted versions of the test sets of CIFAR-10 and ImageNet-100 respectively, comprising of images distorted with $15$ types of corruptions frequently encountered in natural images, broadly grouped into four classes: Noise, Blur, Weather and Digital (\Cref{fig:corruption_images}). Each corruption type has $5$ levels of severity, amounting to a total of $75$ different corruptions. These corruptions do not include Gaussian Blur, thus there is no conflict with Gaussian blurring used during training.

\subsubsection{Metrics} \textit{Clean error} is the top-1 error reported on the standard test set for each of the datasets. Following previous work, we report \textit{mean Corruption Error (mCE)} \cite{hendrycks2018benchmarking} for corruption datasets, computed by averaging the top-1 errors across all corruptions and severity levels on the corruption datasets. 

\subsubsection{Training Details} Unless otherwise mentioned, we use PreAct ResNet-18 architecture \cite{he2016identity}, and train it for 200 epochs on CIFAR-10 and 100 epochs on ImageNet-100. All networks are trained from scratch. The curricula used for training with \texttt{VAC} on the two datasets are noted below. 
\begin{align*}
	& \text{CIFAR-10:} & \quad \mathtt{\{	(13, 2), (27, 1), (160, 0) \}}\\
	& \text{ImageNet-100:}  & \quad \mathtt{\{ (1, 8), (3, 4), (5, 2), (11, 1), (80, 0) \}}
\end{align*}
The designed curricula ensure that the epochs summed over all segments of a curriculum equals the number of epochs for its vanilla training counterpart.

\subsection{Improvement in Common Corruption Robustness} \label{clevr_vs_vanilla}
\begingroup
\setlength{\tabcolsep}{6pt}%
\begin{table*}
	\caption{Clean error and mean Corruption Errors (mCE) of vanilla training compared with \texttt{VAC}. Lower error values are better. Displayed values for CIFAR-10 are the mean error values over 3 repetitions for each entry, expressed as $\%$. For ImageNet-100, we report results from a single run for each experiment.}
	\label{tab:corruptions_vanilla_vs_clewr}
	\centering
	\begin{tabular}{ll|cc|cc}
		\toprule
		& & \multicolumn{2}{c|}{\textbf{Vanilla training}}  & \multicolumn{2}{c}{\textbf{Visual Acuity Curriculum}}\\	
		\textbf{Dataset} & \textbf{Architecture} & \textbf{Clean error ($\downarrow$)} & \textbf{mCE ($\downarrow$)} & \textbf{Clean error ($\downarrow$)} & \textbf{mCE ($\downarrow$)} \\
		\midrule
		\multirow{5}{*}{CIFAR-10}
		& All-Conv \cite{springenberg2014striving}       & 6.13 & 31.19 & 6.77 & \textbf{27.83}  \\
		& ResNet-18 \cite{he2016identity}      & 5.24 & 25.03 & 6.92 & \textbf{18.78}  \\
		& ResNet-56 \cite{he2016identity}      & 6.55  & 28.67 & 10.37 & \textbf{20.37}  \\
		& MobileNet-v2 \cite{sandler2018mobilenetv2} 	  & 14.66 & 33.03 & 16.90 & \textbf{29.60}  \\
		& ConvNeXt \cite{liu2022convnet}      & 7.07 & 17.00 & 7.66 & \textbf{16.39} \\
		\midrule
		\multirow{2}{*}{ImageNet-100}
		& ResNet-18 \cite{he2016identity}      & 13.12 & 53.66 & 15.24 & \textbf{49.23} \\
		& ResNet-50 \cite{he2016identity}      & 10.96 & 51.86 & 15.98 & \textbf{49.03}   \\
		\bottomrule
	\end{tabular}
\end{table*}
\endgroup

\begin{figure}[t]
	\centering
	\includegraphics[width=\linewidth]{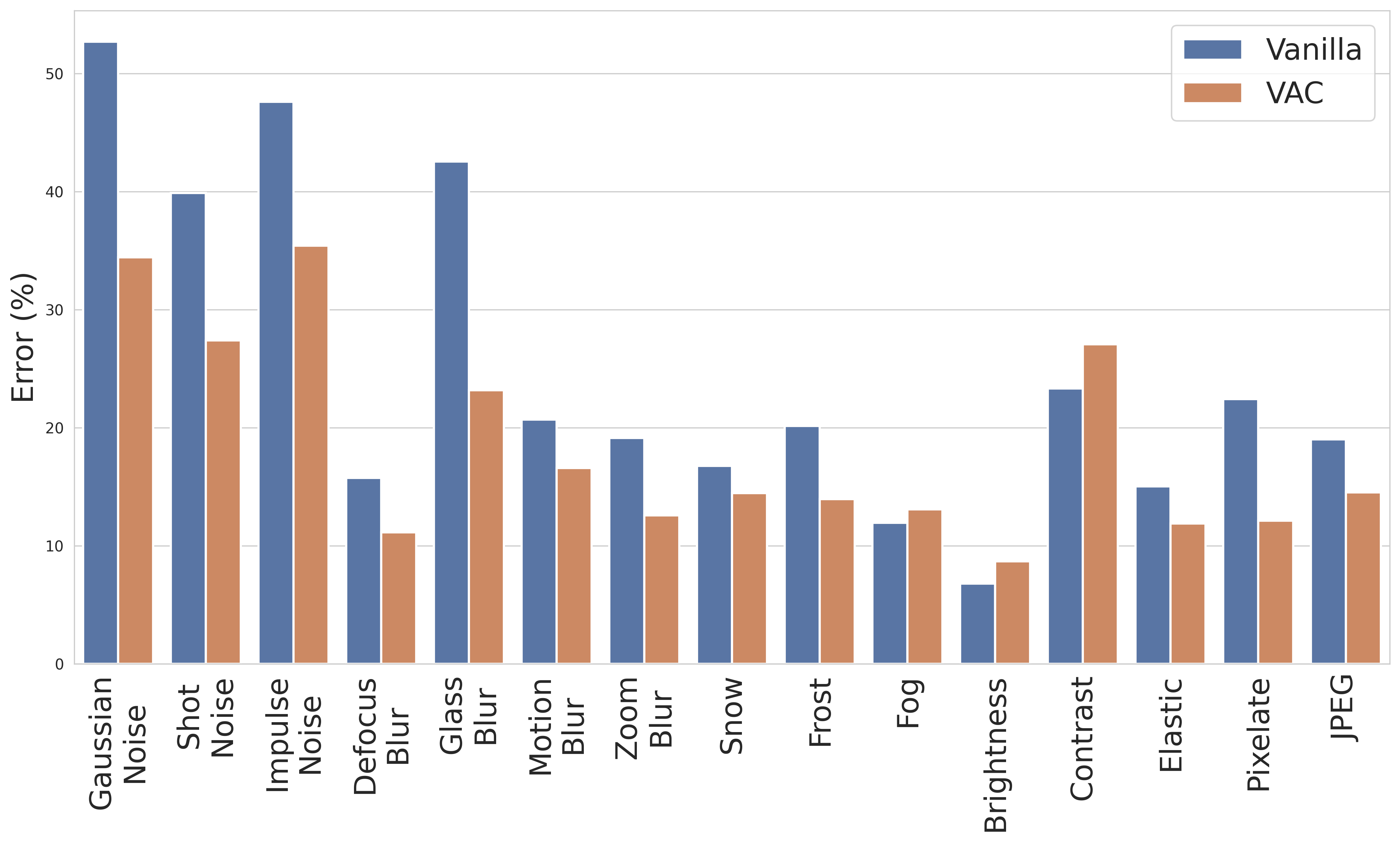}
	\caption{Robustness under data shift: mean Corruption Error across 15 types of image corruptions. Each bar shows the corruption error averaged over 5 severity levels. \texttt{VAC} reduces error (improves accuracy) over most corruption types. }
	\Description{Bar plots comparing the error between Vanilla training (blue) and VAC (orange) across different image corruption types.}
	\label{fig:corruption_types}
\end{figure}

To evaluate the effectiveness of the proposed VAC, we compare it against \textit{vanilla training}, defined as the conventional training of models on clean, high-resolution images from the beginning, without any blur-based curriculum.
As shown in \Cref{tab:corruptions_vanilla_vs_clewr}, \texttt{VAC} consistently improves robustness to common corruptions across different neural network architectures, including both classical models like ResNet-18 and more modern architectures such as ConvNeXt \cite{liu2022convnet}. This suggests that the benefits of VAC are not architecture-specific and can generalize across different model families.

Although \texttt{VAC} leads to a noticeable reduction in mean corruption error (mCE), we observe a small increase in clean (in-domain) error compared to standard training. This trade-off reflects the well-known tension between robustness and clean accuracy. Interestingly, our findings also offer a counterpoint to the Critical Learning Period hypothesis proposed by Achille \emph{et~al.} \cite{achille2018critical}, which suggests that early exposure to degraded input (such as blur) leads to irreversible accuracy loss. While we do observe a slight drop in clean performance, our results indicate that this early-stage information deficit actually promotes more generalizable representations, enabling the network to perform better under distribution shifts. 

To better understand the source of these robustness gains, we present a detailed breakdown of corruption types in \Cref{fig:corruption_types} for a ResNet-18 trained on CIFAR-10. We find that \texttt{VAC}-trained models outperform their vanilla counterparts on most corruption types, including noise-based corruptions (Gaussian noise, impulse noise), weather distortions (snow, frost), and blur-based corruptions (defocus blur, glass blur). These improvements are particularly notable because the model is never explicitly trained on these corruptions. 
%This demonstrates that \texttt{VAC} leads to better out-of-distribution generalization by shaping the early learning dynamics of the network.

However, \texttt{VAC} underperforms on a few corruption categories, most notably brightness and contrast perturbations. We hypothesize that this may be due to the proposed curriculum removing the sparse details which were present in the low-light images. On the other hand, human visual system adapts to low light through an entirely different mechanism not covered in the curriculum. Incorporating such mechanisms, or combining VAC with a photometrically adaptive or multi-scale strategy, could further enhance performance on these challenging corruptions. We leave this as an avenue for future work.

We also note that VAC adds negligible training overhead: training a ResNet-18 on CIFAR-10 with VAC takes 3768s vs. 3729s with vanilla training on an NVIDIA V100 GPU. Importantly, VAC only affects training time; inference speed remains unchanged.

\subsection{Comparison with other Training Strategies}
We now compare the proposed \texttt{VAC} with a selection of established training strategies that aim to improve either generalization, robustness, or learning dynamics. Specifically, we evaluate against FixRes \cite{touvron2019fixing}, Curriculum by Smoothing (CBS) \cite{sinha2020curriculum}, and SuperLoss \cite{castells2020superloss} — three methods with differing motivations and mechanisms.

FixRes \cite{touvron2019fixing} focuses on scale consistency by decoupling resolution from input size and has been shown to improve clean accuracy in high-resolution classification tasks. CBS \cite{sinha2020curriculum} implements curriculum learning by gradually increasing the sharpness of network activations using activation smoothing. SuperLoss \cite{castells2020superloss}, on the other hand, adaptively reweights the training loss based on sample difficulty, helping the network to avoid overfitting to noisy or outlier samples. 

We perform an evaluation of each method across two key robustness benchmarks: corruption robustness using the CIFAR-10-C dataset, and adversarial robustness under a standard PGD (Projected Gradient Descent) \cite{madry2018towards} attack with $\epsilon=2/255$. The results are summarized in \Cref{tab:training_strategies_1}.
\texttt{VAC} outperforms all competing strategies in terms of corruption robustness, achieving the lowest mCE across the board. While FixRes and SuperLoss achieve moderate improvements in clean test accuracy, they show limited or even negative impact on robustness. In some cases, both corruption and adversarial performance degrade relative to vanilla training. This suggests that strategies optimized purely for clean performance may inadvertently compromise the model's resilience to input distribution shifts or adversarial perturbations.

CBS is the only competing method that provides a clear improvement in adversarial robustness, achieving a reduced attack success rate (ASR) under PGD. However, its effectiveness on natural corruptions is limited, and in our experiments, \texttt{VAC} outperforms CBS in terms of mCE. Notably, both \texttt{VAC} and CBS involve a form of curriculum learning, yet the nature of what is being “smoothed” differs fundamentally. CBS smooths internal activations, whereas \texttt{VAC} manipulates the input signal in a biologically motivated way. Our results suggest that manipulating the input signal via Gaussian blur yields more pronounced improvements in robustness to real-world corruptions, possibly because it imposes an inductive bias that encourages the learning of spatially coherent, low-frequency representations.

An interesting finding is that although \texttt{VAC} is not explicitly designed to defend against adversarial attacks, it still improves adversarial robustness over vanilla training, FixRes and SuperLoss. This suggests that the early information deficit enforced by VAC may lead to more stable and noise-tolerant features, which are less susceptible to perturbations, adversarial or otherwise.

\subsection{Compatibility with Data Augmentation}
One of the strengths of the proposed curriculum is its compatibility with a wide range of existing data augmentation strategies. Rather than acting as a replacement, \texttt{VAC} is designed to complement these methods, offering a biologically inspired inductive bias that enhances representation learning. Many recent advances in robustness rely on augmentations that introduce variation in the training distribution, such as spatial manipulations, pixel-level mixing, or adversarial perturbations. \texttt{VAC} can be easily integrated with these approaches to further improve robustness.

To demonstrate compatibility, we combine \texttt{VAC} with five widely-used data augmentation strategies: MixUp \cite{zhang2018mixup}, CutMix \cite{yun2019cutmix}, Adversarial Training \cite{madry2018towards}, RandAugment \cite{cubuk2020randaugment}, and AutoAugment~\cite{cubuk2019autoaugment}. Each of these methods addresses robustness through different mechanisms: MixUp performs convex combinations of images and labels to smooth decision boundaries, CutMix replaces regions of an image with patches from another to enhance spatial generalization, Adversarial Training explicitly incorporates adversarial examples to defend against gradient-based attacks, while RandAugment and AutoAugment operate by learning or randomly sampling from a large search space of augmentation policies to create a more varied and robust training distribution.

Our integration procedure is simple and preserves the design intent of each method. For a given augmentation (e.g., MixUp), we first apply Gaussian blur to the input image according to the current \texttt{VAC} curriculum stage. The blurred images are then passed through the selected augmentation pipeline. This ordering reflects the natural perceptual sequence in humans, where low-resolution global structure is processed first, followed by exposure to diverse contextual variations.

As shown in \Cref{tab:data_augmentation}, combining \texttt{VAC} with these augmentation strategies leads to further improvements in both common corruption robustness and adversarial robustness. These results support our hypothesis that \texttt{VAC} introduces a complementary inductive bias that improves the effectiveness of existing data augmentations, and can serve as a general-purpose module in robustness-focused training pipelines, particularly when combined with augmentation strategies that introduce orthogonal forms of variability.

\begingroup
\setlength{\tabcolsep}{10pt}%
\begin{table}[t]
	\caption{Comparison of \texttt{VAC} with other training strategies, evaluated for a ResNet-18 network trained on CIFAR-10. We report clean error and common corruption errors ($\%$) and Attack Success Rate (ASR) for PGD adversarial attack.}
	\label{tab:training_strategies_1}
	\centering
	\begin{tabular}{lccc}
		\toprule
		{\textbf{Method}}  & {\textbf{Clean error }} & {\textbf{Mean CE}} & \textbf{ASR} \\
		& ($\downarrow$) & ($\downarrow$) & ($\downarrow$) \\
		\midrule
		Vanilla Training			& 5.43 & 24.60 & 54.31 \\
		FixRes \cite{touvron2019fixing}	& 5.49 & 27.21 & 59.24 \\
		CBS	\cite{sinha2020curriculum} 						& 5.79 & 27.87 & \textbf{46.23}\\
		SuperLoss \cite{castells2020superloss}					& 5.74 & 27.07 & 58.44 \\
		\texttt{VAC} (Proposed)					& 6.63 & \textbf{17.58} & 50.37\\
		\bottomrule		
	\end{tabular}
\end{table}
\endgroup

\begingroup
\setlength{\tabcolsep}{10pt}%
\begin{table}[t]
	\caption{Performance of \texttt{VAC} combined with data augmentation techniques on a ResNet-18 trained on CIFAR-10. We report clean error, common corruption error, and Attack Success Rate (ASR) under a PGD adversarial attack.}
	\label{tab:data_augmentation}
	\centering
	\begin{tabular}{lccc}
		\toprule
		{\textbf{Method}}  & {\textbf{Clean}} & {\textbf{Mean CE}} & \textbf{ASR} \\
		& \textbf{error} ($\downarrow$) & ($\downarrow$) & ($\downarrow$) \\
		\midrule
		MixUp \cite{zhang2018mixup} & 4.40 & 19.00 & 68.01 \\
		MixUp + \texttt{VAC} & 5.68 & \textbf{18.66} & \textbf{57.89} \\
		\midrule
		CutMix \cite{yun2019cutmix} & 4.14 & 28.90 & 73.55 \\
		CutMix + \texttt{VAC} & 4.87 & \textbf{21.99} & \textbf{69.39}  \\
		\midrule
		$\ell_2$ adversarial \cite{madry2018towards} & 6.24 & 16.29 & \textbf{21.01} \\
		$\ell_2$ adversarial + \texttt{VAC} & 9.01 & \textbf{15.82} & 21.36 \\
		\midrule
		RandAugment \cite{cubuk2020randaugment} & 4.97 & 18.77 & 52.06 \\
		RangAugment + \texttt{VAC} & 6.08 & \textbf{14.59} & \textbf{49.04}  \\
		\midrule
		AutoAugment \cite{cubuk2019autoaugment} & 4.78 & 18.21 & 52.82 \\
		AutoAugment + \texttt{VAC} & 6.07 & \textbf{15.04} & \textbf{45.79}  \\
		\bottomrule		
	\end{tabular}
\end{table}
\endgroup

\begin{figure*}[thbp]
	\centering
	\includegraphics*[width=\linewidth]{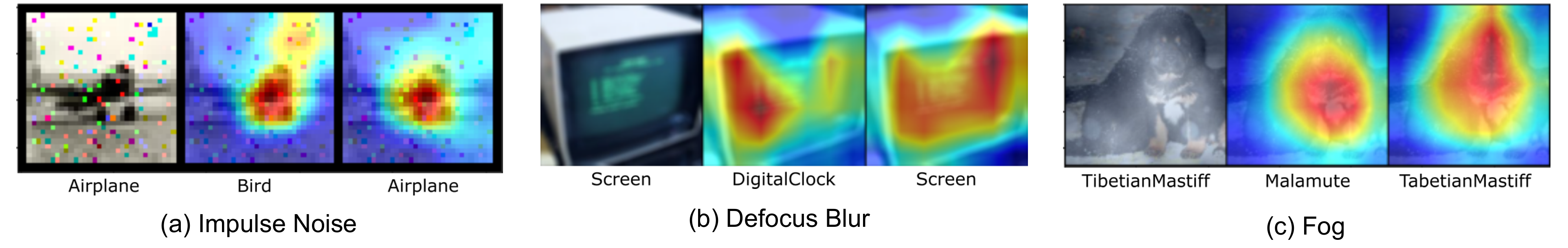}
	\caption{Grad-CAM visualizations: (a) shows image from CIFAR-10, and (b) and (c) from ImageNet-100. Each set shows results for a particular image corruption. First column in a set is the corrupted image, second and third columns are the Grad-CAMs created for Vanilla and \texttt{VAC} trained networks respectively. Original and predicted labels are depicted below the images. The discriminative regions in the CAM output are larger and more pronounced for \texttt{VAC}.}
	\Description{An image of an airplane with impulse noise, a TV screen with defocus blur, and a Tibetan Mastiff with fog. For each image, the corresponding Grad CAM visualizations for Vanilla and VAC trained networks are shown; red color represents regions of high importance, and blue shows regions with low importance.}
	\label{fig:gradcam}
	\vspace{2em}
\end{figure*}

\subsection{Ablation Study}
\label{ablation_studies}

The proposed curriculum is motivated by the developmental trajectory of the human visual system, which gradually improves visual acuity over time. While this inspiration provides a compelling biological basis, our final curriculum design involves several heuristics and practical choices. In this section, we perform an ablation study to assess the effect of these design decisions.

\subsubsection{Curriculum variants}
Firstly, e compare \texttt{VAC} with several curriculum variants and blur-based baselines in \Cref{tab:ablation}:
\begin{enumerate*}[label=(\roman*),leftmargin=0.5cm]
	\item \textbf{Linear Curriculum:} This tests whether an evenly spaced progression is as effective as our proposed exponential decay. The schedule is defined as $\mathtt{\{(20,2), (20,1), (160,0)\}}$, where the blur level decreases in equal-duration steps. 
	\item \textbf{Inverse Curriculum:} Defined as $\mathtt{\{(13,0), (27,1), (160,2)\}}$, this curriculum starts with clean images and increases blur over time. It serves as a counter-hypothesis to test whether early access to detailed features provides any robustness benefit.
	\item \textbf{Continuous Curriculum:} Here, we fix the blur level at $\sigma=2$ and gradually reduce the proportion of blurred images in each mini-batch over time, rather than staging the transition in discrete segments. This tests whether blur quantity, rather than sharp transition boundaries, is the critical factor.
	\item \textbf{Steep Curriculum (No Replay):} A simplified schedule defined as $\mathtt{\{(20,2), (180,0)\}}$ that makes a sharp transition from high blur to clean inputs without incorporating replay. This variant is used to isolate the effect of the replay mechanism introduced in \texttt{VAC}.
	\item \textbf{Constant Blur:} We test two cases: one where $100\%$ of inputs in every batch are blurred using $\sigma=2$, and another with $20\%$ blurred inputs. These serve as baselines for conventional blur-based augmentation without curriculum structure.
\end{enumerate*}

\begingroup
\setlength{\tabcolsep}{10pt}%
\begin{table}[t]
	\caption{Ablation study. Clean and corruption errors ($\%$) on ResNet-18 trained on CIFAR-10, for alternative training regimens.}
	\label{tab:ablation}
	\centering
	\begin{tabular}{lcc}
		\toprule
		\textbf{Method}  & \textbf{Clean error} ($\downarrow$) & \textbf{Mean CE} ($\downarrow$) \\
		\midrule
		\texttt{VAC} (Proposed)		& 6.63 & \textbf{17.58} \\
		Linear curriculum 		& 7.07 & 17.99 \\
		Inverse curriculum 		& 14.95 & 35.98 \\
		Continuous curriculum 	& 6.33 & 20.28 \\
		Steep (no replay)		& 20.21 & 23.52 \\
		Constant blur $100\%$ 	& 67.67 & 68.16 \\
		Constant blur $20\%$ 	& \textbf{6.18} & 18.02 \\
		\bottomrule
		
	\end{tabular}
\end{table}
\endgroup

The results indicate that the choice and scheduling of blur exposure significantly impact both clean and corruption robustness. The \textit{inverse curriculum}, which delays the introduction of blur until later training epochs, yields the worst corruption and clean error, confirming that late-stage deficits harm representational fidelity. Similarly, applying high blur uniformly throughout training (\textit{constant blur at 100\%}) leads to degraded performance, suggesting that excessive visual suppression limits the model’s ability to learn fine-grained discriminative features.
The \textit{steep curriculum} variant, which drops blur abruptly after the initial phase without replay, suffers from markedly high clean error. This supports our hypothesis that catastrophic forgetting erodes representations learned during the initial blurred phase unless earlier inputs are periodically replayed. The sharp drop in performance underlines the necessity of our replay mechanism.
While \textit{linear} and \textit{continuous} curricula provide reasonable trade-offs, they fall short of the performance achieved by the proposed schedule. Notably, the proposed exponential decay schedule with replay achieves the best balance between clean accuracy and corruption robustness.

This ablation study thus validates several key aspects of our design: the importance of starting with high blur, reducing blur progressively rather than abruptly, and reinforcing earlier blur levels via replay. These components collectively contribute to the effectiveness of \texttt{VAC} in learning robust representations.

\subsubsection{Deficit period} The choice of 20\% deficit epochs ($N_\textrm{def} = \lfloor{N/5}\rfloor$ in \Cref{alg1}) is based on empirical evaluation. Increasing the deficit in the range of 5\% to 40\% reduces corruption error (19\% to 17.3\%) but increases clean error to 9.02\%, with diminishing returns after 30\%. The 20\% deficit balances clean and corruption errors well. 

\subsubsection{Guidance on choosing $\sigma_{max}$} The choice of $\sigma_{max}$ depends on input resolution: we use $\sigma_{max}$ = 2 for 32×32 and $\sigma_{max}$ = 8 for 224×224 images. Larger $\sigma_{max}$ increases initial blur, impacting clean accuracy more than mCE. For example, on ImageNet-100 with ResNet-18, increasing $\sigma_{max}$ from 4 to 8 to 16 raises clean error from 14.18\% to 15.24\% and 17.48\%. We recommend tuning $\sigma_{max}$ via grid search based on image size to balance robustness and accuracy.

\subsubsection{Deficit vs. Data Augmentation} \label{deficit_vs_aug}
The proposed method creates a sensory deficit by exposing the network to \textit{only} low resolution images in the early epochs. A close alternative to the proposed curriculum is the commonly used blur-based Data Augmentation, where blurred images are \textit{intermixed} with clean images over the entire training period, represented by the \textit{constant blur 20\%} method in \Cref{tab:ablation}. Constant blur 20\% achieves lower clean error compared to \texttt{VAC}, which can be attributed to the fact that there is no deficit enforced in the initial epochs. However, \texttt{VAC} achieves lower corruption error, suggesting the benefits of stimulus deficit in the early epochs compared to regular data augmentation with blurred images. 

\subsection{Grad-CAM Visualizations}

To better understand the qualitative differences in feature representations learned by models trained with the Visual Acuity Curriculum (\texttt{VAC}), we employ Grad-CAM \cite{selvaraju2017grad} to visualize class-discriminative regions in the input images. \Cref{fig:gradcam} presents comparisons between models trained with \texttt{VAC} and those trained via standard (vanilla) procedures on examples drawn from the CIFAR-10 and ImageNet-100 datasets.

We observe that, under a variety of image degradations, such as blur, noise, or fog, the Grad-CAM maps from \texttt{VAC}-trained networks tend to focus on broader and more semantically meaningful regions of the image. In contrast, vanilla-trained models often localize on narrow or fragmented regions, many of which do not clearly correspond to object-specific features.

These visualizations support our hypothesis that the early-stage information deficit imposed by \texttt{VAC} encourages the model to rely on coarse, low-frequency features and global structural cues. As a result, the network learns to extract more spatially coherent and resilient representations that generalize better under distribution shifts. The broader activation patterns suggest that \texttt{VAC}-trained models are less prone to overfitting on fine-grained, potentially spurious details in the training data.

\section{Conclusion and Future Work}

We proposed a training curriculum for neural networks inspired by the gradual development of visual acuity in human infants. Our experiments show that this biologically motivated approach enhances the robustness of CNNs against a wide range of unseen image corruptions and distribution shifts. Rather than replacing existing robustness techniques, our method is complementary, offering a new inductive bias grounded in human vision. By simulating an early-stage sensory deficit and gradually increasing input detail during training, we aim to bridge a small but meaningful gap between biological and artificial visual learning. Our findings suggest that such developmental priors can guide networks toward more generalizable and resilient representations.

This work opens several promising avenues for future research. Other aspects of human visual development, such as the progressive emergence of color perception or binocular depth processing, may also provide useful principles for designing training curricula. Another exciting direction is to study how the proposed curriculum interacts with vision transformers (ViTs), which, unlike CNNs, rely on self-attention mechanisms to model global context. 
%It remains an open question whether a progressive visual acuity schedule would benefit or interfere with transformer learning dynamics. 
Investigating whether gradually increasing input resolution improves ViT robustness and sample efficiency could lead to broader insights into curriculum design for modern architectures.

\begin{acks} 
	The authors would like to thank Soumen Basu, Ashutosh Agarwal, Devesh Pant and Gaurav Talebailkar for their help in carrying out some of the experiments. 
\end{acks}

\bibliographystyle{ACM-Reference-Format}
\bibliography{hvd_ref}

\end{document}